\documentclass[sigconf]{acmart}

\AtBeginDocument{%
  \providecommand\BibTeX{{%
    \normalfont B\kern-0.5em{\scshape i\kern-0.25em b}\kern-0.8em\TeX}}}

\copyrightyear{2023} 
\acmYear{2023} 
\setcopyright{acmlicensed}
\acmConference[WWW '23]{Proceedings of the ACM Web Conference 2023}{April 30-May 4, 2023}{Austin, TX, USA}
\acmBooktitle{Proceedings of the ACM Web Conference 2023 (WWW '23), April 30-May 4, 2023, Austin, TX, USA}
\acmPrice{15.00}
\acmDOI{10.1145/3543507.3583300}
\acmISBN{978-1-4503-9416-1/23/04}

\usepackage[utf8]{inputenc}
\usepackage{xcolor}
\usepackage{balance}
\usepackage{mathtools}
\usepackage{url}
\usepackage{amsfonts}
\usepackage{algorithmic}
\usepackage{textcomp}
\usepackage{algorithm}
\usepackage{multirow}
\usepackage{color}
\usepackage{subcaption}
\usepackage{enumitem}

\newcommand{\m}{\textsf{KHAN}}
\newcommand{\han}{\textsf{HAN}}
\newcommand{\ke}{\textsf{KE}}
\newcommand{\codeurl}{\url{https://github.com/yy-ko/khan-www23}}

\begin{document}
\title[{\m}: Knowledge-Aware Hierarchical Attention Networks for Accurate Political Stance Prediction]{{\m}: Knowledge-Aware Hierarchical Attention Networks\\for Accurate Political Stance Prediction}





\author{Yunyong Ko}
\email{yyko@illinois.edu}
\affiliation{%
  \institution{University of Illinois at Urbana-Champaign, IL, USA}
  \city{} 
  \country{}
}

\author{Seongeun Ryu}
\email{ryuseong@hanyang.ac.kr}
\affiliation{%
  \institution{Hanyang University}
  \city{Seoul}
  \country{Republic of Korea}
}

\author{Soeun Han}
\email{sosilver@hanyang.ac.kr}
\affiliation{%
  \institution{Hanyang University}
  \city{Seoul}
  \country{Republic of Korea}
}

\author{Youngseung Jeon}
\email{ysj@g.ucla.edu}
\affiliation{%
  \institution{University of California}
  \city{Los Angeles, CA}
  \country{USA}
}

\author{Jaehoon Kim}
\email{jaehoonkimm@hanyang.ac.kr}
\affiliation{%
  \institution{Hanyang University}
  \city{Seoul}
  \country{Republic of Korea}
}

\author{Sohyun Park}
\email{sally5004@ajou.ac.kr}
\affiliation{%
  \institution{Ajou University}
  \city{Suwon}
  \country{Republic of Korea}
}

\author{Kyungsik Han}
\email{kyungsikhan@hanyang.ac.kr}
\affiliation{%
  \institution{Hanyang University}
  \city{Seoul}
  \country{Republic of Korea}
}

\author{Hanghang Tong}
\email{htong@illinois.edu}
\affiliation{%
  \institution{University of Illinois at Urbana-Champaign, IL, USA}
  \city{} 
  \country{}
}

\author{Sang-Wook Kim}
\authornote{Corresponding author}
\email{wook@hanyang.ac.kr}
\affiliation{%
  \institution{Hanyang University}
  \city{Seoul}
  \country{Republic of Korea}
}

\renewcommand{\shortauthors}{Y. Ko et al.}

\begin{abstract}
The political stance prediction for news articles has been widely studied to mitigate the \textit{echo chamber} effect 
-- people fall into their thoughts and reinforce their pre-existing beliefs.
The previous works for the political stance problem focus on (1) identifying political factors that could reflect the political stance of a news article and (2) capturing those factors effectively.
Despite their empirical successes,
they are not sufficiently justified in terms of how effective their identified factors are in the political stance prediction.
Motivated by this, in this work,
we conduct a user study to investigate important factors in political stance prediction, 
and observe that the \textit{context} and \textit{tone} of a news article (\textit{implicit}) and \textit{external knowledge} for real-world entities appearing in the article (\textit{explicit}) are important in determining its political stance. 
Based on this observation, 
we propose a novel knowledge-aware approach to political stance prediction (\textbf{{\m}}), 
employing (1) hierarchical attention networks ({\han}) to learn the relationships among words and sentences in three different levels and 
(2) knowledge encoding ({\ke}) to incorporate external knowledge for real-world entities into the process of political stance prediction.
Also, to take into account the subtle and important difference between opposite political stances, 
we build two independent political knowledge graphs (KG) (i.e., \textsf{KG-lib} and \textsf{KG-con}) by ourselves and learn to fuse the different political knowledge.
Through extensive evaluations on three real-world datasets,
we demonstrate the superiority of {\m} in terms of (1) accuracy, (2) efficiency, and (3) effectiveness.
\end{abstract}



\maketitle

\section{Introduction}
\label{sec:intro}

With the prevalence of web-based news platforms, people are having more chances to obtain a variety of high-quality information about social and political issues.
In general, people tend to prefer news contents which have similar political stances to them~\cite{kruglanski1996motivated}.
For example, people who have a conservative stance often prefer news articles from conservative news media such as Fox and Breibart, 
while those with a liberal stance might prefer articles from liberal news media such as CNN, New York Times, and Washington Post.
As this tendency becomes intensified, however, people could be trapped in their own opinions and reinforce their pre-existing beliefs by limiting exposure to other news articles having different opinions. 
This is called the \textit{echo chamber} effect~\cite{shu2017fake}, a fundamental reason that leads to serious social polarization by hindering positive and active communications among people~\cite{flaxman2016filter,garimella2018political,zannettou2018gab,cota2019quantifying,cinelli2021echo,han2022defend}.
In addition, machine learning (ML)-based recommendation systems have been exacerbating this problem since their goal is to recommend news articles that are likely to be preferred by users~\cite{karimi2018news,garrett2011resisting,bakshy2015exposure,quattrociocchi2016echo}.

\begin{figure}[t]
\centering
\includegraphics[width=1\columnwidth]{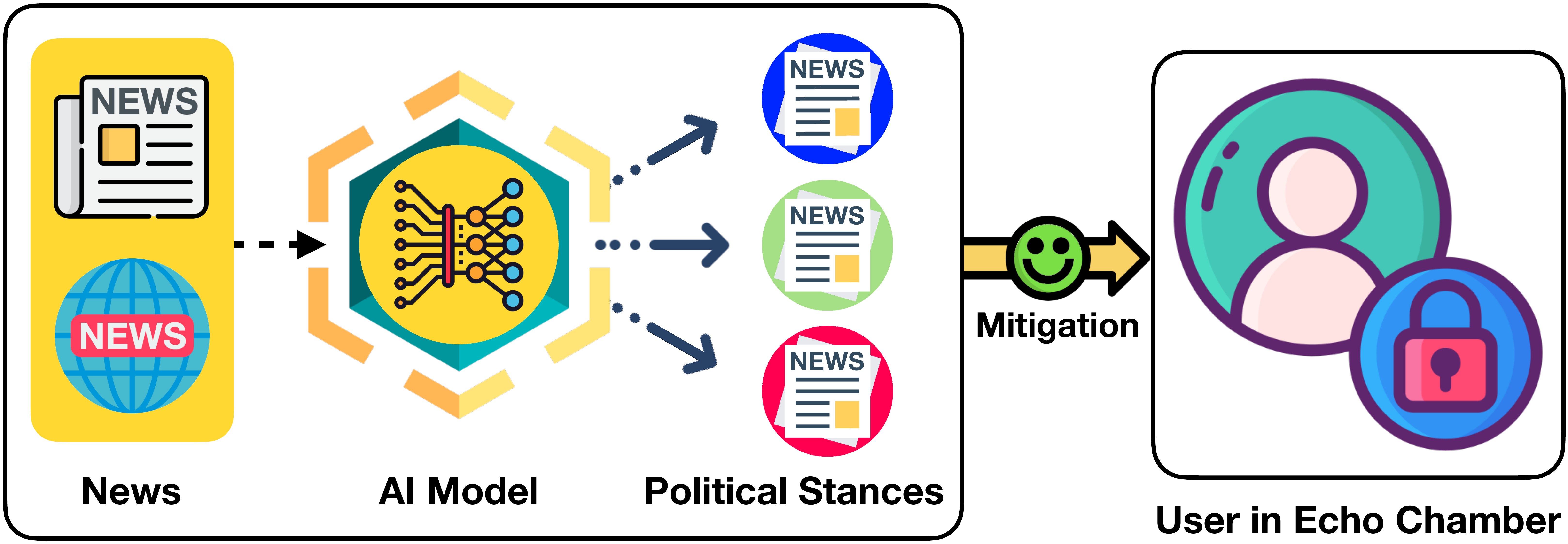}
\caption{Accurate provision of diverse political stances to mitigate the echo chamber effect.}
\vspace{-5mm}
\label{fig:breaking-echo-chamber}
\end{figure}

Many studies have been proposed to mitigate the echo chamber effect~\cite{bail2018exposure,liao2014can,munson2013encouraging,gillani2018me,kriplean2012supporting,liao2015all,nelimarkka2019re}.
They mainly focus on exposing \textit{diverse} opinions (e.g., news articles with different stances) to people in order to prevent them from falling into their existing beliefs too much.
In computer science, for example, 
many groups of researchers have studied the \textit{diversification} of news recommendation systems in order to provide news articles
not only interesting but also fresh to users~\cite{ziegler2005improving,lunardi2019representing,hassan2019trust,zheng2021dgcn}.
Through this effort, people could view a given issue from various perspectives and understand how rational or biased they are, 
thereby leading to the mitigation of the echo chamber effect~\cite{dahlberg2001internet}.
Based on this understanding, from the technical aspect, 
it is critical to accurately predict the political stance of a given news article (i.e., \textit{political stance prediction}), which this paper focuses on.
This is because the accurate provision of news articles with different political stances would not only (1) have people experience diverse political stances
but also (2) allow researchers to investigate ways to support news consumption in a balanced standpoint from people's experiences and behaviors~\cite{liao2014can}.
Although there has been much effort to predict political stances of news articles by domain experts,
it encompasses a manual process of determining political stances that requires a great amount of time and effort and may be also influenced by human biases~\cite{jeon2021chamberbreaker}.

For this political stance prediction, 
existing language models such as GloVe~\cite{pennington2014glove}, BERT~\cite{kenton2019bert}, and RoBERTa~\cite{liu2019roberta} can be applied.
However, it is reported that a single application of language models empirically failed to achieve high accuracy in the political stance prediction because they are not designed specially for predicting the political stance of a news article~\cite{li2021using,feng2021kgap,zhang2022kcd}.
Recently, in order to overcome the limitation of general language models, 
deep neural networks (DNN) models for political stance prediction have been proposed~\cite{li2019encoding,li2021using,feng2021kgap,zhang2022kcd}.
They identify social, linguistic, and political factors reflecting the political stance of a news article and design new model architectures to effectively capture the identified factors.
Although these DNN models empirically achieved higher accuracies than existing language models, 
they are not sufficiently justified in terms of why and how effective their identified factors are in predicting the political stances of news articles.

From this motivation, 
we first conducted a user study in order to investigate what factors real-world users consider and how important the factors are in determining the political stances of news articles.
We provided six articles with different political stances and carefully-chosen factors to 136 respondents via Amazon Mechanical Turk. 
We asked them to respond how important each factor is in their decision with a scale [1:(not at all)-5:(very much)]\footnote{The details of the user study are described in Appendix~\ref{sec:appendix-userstudy}.}.
Through the user study, we observe that 
the "context" of a new article is the most important factor in deciding its political stance, followed by keyword, person, tone, and frequently used word.
This result gives us an important lesson: 
it is crucial (1) \textit{to learn the relationships among words/sentences to capture the context and tone of a news article}, which is \textit{implicitly} reflected in the article, 
and (2) \textit{to understand the interpretation and sentiment to real-world entities (e.g., keyword and person)}, which \textit{explicitly} appear in a news article.

Towards reflecting both the explicit and implicit factors, in this paper,
we propose a novel approach to accurate political stance prediction, 
named \textbf{\underline{K}}nowledge-aware \textbf{\underline{H}}ierarchical \textbf{\underline{A}}ttention \textbf{\underline{N}}etworks (\textbf{{\m}}).
{\m} consists of two key components: 
(1) hierarchical attention networks ({\han}) to learn the relationships among words/sentences in a news article with 3-level hierarchy (i.e., word-level, sentence-level, and title-level), rather than learning the entire article itself,
and (2) knowledge encoding ({\ke}) to incorporate both common and political knowledge for real-world entities, necessary for understanding a news article, into the process of predicting the political stance of a news article.
Regarding political knowledge, the interpretation and sentiment to even the same entities can be different depending on its political stance~\cite{gillani2018me,liao2014can, munson2013encouraging}.
To address the subtle but important difference,
we construct two knowledge graphs (KG) with different political stances, \textsf{KG-lib} and \textsf{KG-con}, 
and design {\ke} to learn to fuse the information extracted from the political knowledge graphs.
To our best knowledge, this is the first work that leverages both common and political knowledges separately, 
further reflecting the different political knowledges.

The main contributions of this work are as follows.

\begin{itemize}[leftmargin=10pt]
    \item \textbf{Datasets}: To reflect the different political knowledge of each entity, we build two political knowledge graphs, \textsf{KG-lib} and \textsf{KG-con}.
    Also, for extensive evaluation, we construct a large-scale political news datatset, \textsf{AllSides-L}, much larger ($48\times$) than the existing largest political news article dataset.\footnote{The data construction details (\textsf{KG-lib}, \textsf{KG-con}, \textsf{AllSides-L}) are provided in Appendix~\ref{sec:appendix-data}.}
    
    \item \textbf{Algorithm}: We propose a novel approach to accurate political stance prediction ({\m}), employing (1) hierarchical attention networks ({\han}) and (2) knowledge encoding ({\ke}) to effectively capture both explicit and implicit factors of a news article.
    
    \item \textbf{Evaluation}: Via extensive experiments, we demonstrate that (1) (\textit{accuracy}) {\m} consistently achieves higher accuracies than all competing methods (up to 5.92\% higher than the state-of-the-art method), 
    (2) (\textit{efficiency}) {\m} converges within comparable training time/epochs, 
    and (3) (\textit{effectiveness}) each of the main components of {\m} is effective in political stance prediction.
\end{itemize}

\noindent
For reproducibility, we have released the code of {\m} and the datasets at {\codeurl}.

\section{Related Work}\label{sec:related}

\noindent
\textbf{Language models.}
The political stance prediction can be seen as a special case of document classification. 
Thus, general language models~\cite{mikolov2013efficient,pennington2014glove,peters-etal-2018-deep,kenton2019bert,liu2019roberta,radford2018improving,lan2019albert,clark2020electra}, 
which aim at learning to represent words into embedding vectors, could be applied to the political stance prediction problem.
Word2Vec~\cite{mikolov2013efficient} is a traditional language model that learns word embeddings based on the similarity between words to preserve local context.
However, Word2Vec does not reflect the global context in the entire document~\cite{chen2017efficient}. 
To address the limitation of Word2Vec, 
GloVe~\cite{pennington2014glove} utilizes not only the local relationships among words but also global information of a given document. 
ELMo~\cite{peters-etal-2018-deep} aims to learn the meanings of words that can be varying depending on the context of a given document, using the pre-trained Bi-LSTM model as contextual information.
In addition, \textit{task-agnostic} language models, pre-trained based on large-scale data, can be applied to the political stance prediction problem by fine-tuning them on political news article data.
BERT~\cite{kenton2019bert}, composed of multiple transformer encoders~\cite{vaswani2017attention}, learns the context of given text in a \textit{bidirectional} way with the masked language model (MLM) and the next sentence prediction (NSP).
RoBERTa (Robustly optimized BERT)~\cite{liu2019roberta}, pre-trained on much more training data than BERT, proposes a dynamic masking technique to improve MLM of BERT, achieving higher accuracy than that of BERT.
These language models, however, have a limitation in capturing the political characteristics of news articles since they are not designed specially for predicting the political stance of a news article~\cite{li2021using,feng2021kgap,zhang2022kcd}.

\vspace{2mm}
\noindent
\textbf{Political stance prediction models.}
To overcome the limitation of language models, 
many researchers have studied methods that consider both text and additional information (e.g., social, linguistic, and political information), which could be useful in political stance prediction~\cite{li2019encoding,li2021using,feng2021kgap,zhang2022kcd,sun2018stance,li2021mean}.
HLSTM (Hierarchical LSTM)~\cite{li2019encoding} encodes a news article into an embedding vector using hierarchical LSTM models and it additionally utilizes social context information (e.g., how the news article is spread across users)~\cite{ko2016accurate,ko2018efficient,ko2018influence} that could reflect the political stance of a news article.
Similarly, MAN~\cite{li2021using} learns the relationships among words in news articles using multi-head attention networks and considers social and linguistic information necessary to understand the political context of a news article.
More recently, knowledge graph (KG) based approaches to political stance prediction have been proposed~\cite{feng2021kgap, zhang2022kcd}.
A KG-based approach constructs a knowledge graph based on political news articles and uses the political knowledge extracted from the KG as additional information.
KGAP (Knowledge Graph Augmented Political perspective detection)~\cite{feng2021kgap} represents a given news article as a 4-layer graph (word, sentence, paragraph, and article nodes), 
and then, it injects the knowledge information to each word node and applies graph neural networks (e.g., R-GCN~\cite{schlichtkrull2018modeling}) to the article graph.
KCD (Knowledge walks and textual Cues enhanced political perspective Detection)~\cite{zhang2022kcd}, a state-of-the-art political stance prediction model, 
generates political knowledge walks via performing random walks in the political KG (like simulating the process of human reasoning) and combines the political knowledge walks with the text of a news article using multi-head attention layers.
\section{The Proposed Method: {\m}}\label{sec:proposed}

In this section, we present a novel approach to political stance prediction, \textbf{\underline{K}}nowledge-aware \textbf{\underline{H}}ierarchical \textbf{\underline{A}}ttention \textbf{\underline{N}}etworks (\textbf{{\m}}).
First, we describe the notations and the problem definition. 
Then, we present two main components of {\m}: hierarchical attention networks ({\han}) and knowledge encoding ({\ke}).

\subsection{Problem Definition}\label{sec:proposed-problem}
The notations used in this paper are described in Table~\ref{table:notations}.
{\m} manages two types of embeddings in order to hierarchically learn the relationships among words and the relationships among sentences in a news article: 
word embeddings ($W=\{w_1, w_2, ..., w_N\}$) and sentence embeddings ($S=\{s_1, s_2, ..., s_{l}\}$),
where $w_i$ ($s_i$) represents the $d$-dimensional word (sentence) embedding of the $i^{th}$ word (sentence).
In addition to word and sentence embeddings, 
{\m} also manages external knowledge embeddings to incorporate common and political knowledge in political stance prediction: 
common and political knowledge embeddings ($K^{com}$, $K^{lib}$, and $K^{con}$).

\vspace{1mm}
\noindent
\textit{\textbf{Political stance prediction}.}
This work aims to solve the \textit{political stance prediction} problem:  
given a news article $a$, predict its political stance (e.g., [1, 5], where 1 indicates `left' and 5 indicates `right').
This problem, a typical supervised learning task, can be defined as follows from the optimization perspective:
\begin{equation}
\min_{w \in \mathbb{R}} \frac{1}{|A|} \sum_{a \in A} F(X,a) + || X ||^2,
\label{eq:objective}
\end{equation}
where $X$ is the set of learnable parameters, $A$ is a given news article dataset,
and $F(X, a)$ is the loss function of the parameters $X$ given a news article $a$.
As the loss function, $F(\cdot)$, we adopt cross-entropy loss with the $L_2$-regularization term.

Based on this problem definition, we optimize the learnable parameters $X$ in an \textit{end-to-end} way.
More specifically, to solve the problem represented in Eq~\ref{eq:objective}, we consider SGD as an optimization algorithm.
Let $X_t$ be the model parameters at iteration $t$. 
Then, $X_t$ is optimized iteratively by the following rule: $X_{t+1} = X_t-\eta \cdot \nabla F(X_t, a)$,
where $\eta$ is the user-defined learning rate.

\begin{table}[t]
\caption{Notations and their descriptions.}
\vspace{-3mm}
\setlength\tabcolsep{6pt}
\begin{tabular}{cl}
\toprule
 \textbf{Notation} & \textbf{Description}\\
\midrule
$W$ & a set of word embeddings \\
$S$ & a set of sentence embeddings \\
$T$ & the title embedding \\
$w_i,s_j$ & $i^{th}$ word embedding, $j^{th}$ sentence embedding \\
$d$ & the embedding dimensionality \\
\midrule
$N$ & the total number of words in a dataset \\
$n$ & the maximum number of words in a sentence \\
$l$ & the maximum number of sentences in an article \\
\midrule
$K^{com}$ & a set of common knowledge embeddings \\
$K^{lib}$ & a set of liberal knowledge embeddings  \\
$K^{con}$ & a set of conservative knowledge embeddings  \\
$\alpha, \beta$ & common and political knowledge factors \\
\midrule
$A, a$ & news article dataset and a news article \\
$X$ & a set of learnable parameters \\
$F(\cdot)$ & loss function (i.e., cross-entropy loss) \\
$\eta$ & user-defined learning rate \\

\bottomrule
\end{tabular}
\label{table:notations}
\end{table}

\begin{figure*}[t]
\centering
\includegraphics[width=0.95\textwidth]{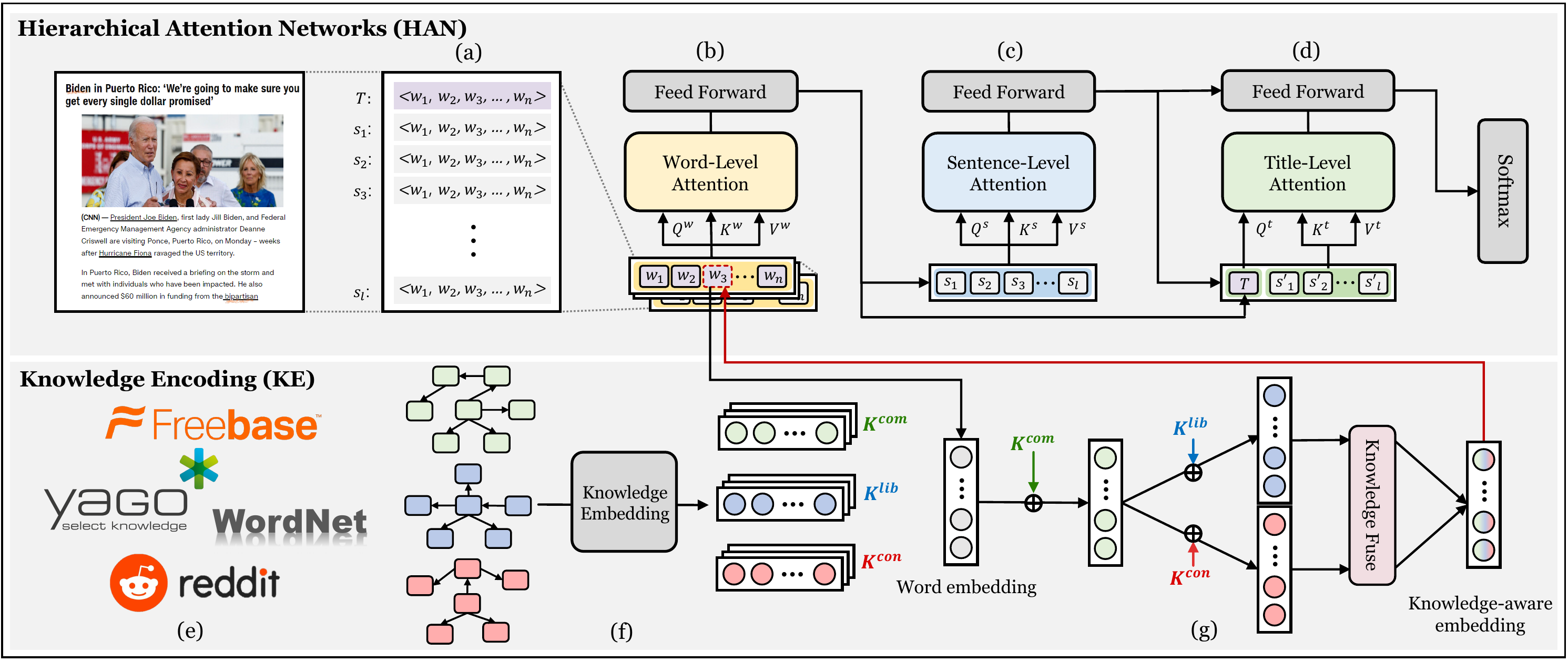}
\vspace{-3mm}
\caption{The overview of {\m}: hierarchical attention networks (upper) and knowledge encoding (lower).}
\vspace{-2mm}
\label{fig:overview}
\end{figure*}

\subsection{Hierarchical Attention Networks ({\han})}\label{sec:proposed-han}

As shown in Table~\ref{table:userstudy}, the context and tone of a news article are important factors in predicting the political stance of a news article.
These factors, however, do not appear in the article explicitly; instead, they are \textit{implicitly} reflected in the overall article.
Due to their implicit characteristics, 
it is very challenging to capture the context and tone of a news article.
To address this challenge, in this section, 
we propose a \textbf{\underline{H}}ierarchical \textbf{\underline{A}}ttention \textbf{\underline{N}}etworks (\textbf{{\han}}) that learns (1) the relationships among words, (2) the relationships among sentences, and (3) the relationships between the title and sentences (i.e., 3-level hierarchy), for effectively capturing the implicit factors (i.e., context and tone) from news articles.
Through the 3-level hierarchy, 
{\han} is able to sequentially capture (1) the local context in each sentence, (2) the global context in an article, and (3) the key idea included in the title.

{\han} consists of four layers as illustrated in Figures~\ref{fig:overview}(a)-(d): 
(a) a pre-processing layer;
(b) a word-level attention layer;
(c) a sentence-level attention layer;
and (d) a title-level attention layer. 
Now, we describe the four layers and their advantages and challenges.

\vspace{1mm}
\noindent
\textbf{Pre-processing layer.}
In general, a news article data is not structured and just represented as a sequence of words (or symbols).
Thus, to apply our hierarchical attention networks to a news article, 
we need to pre-process an input article to represent as a structured form.
We pre-process a given news article as follows: Given a news article $a$, 
we transform the input article into a set of sentences based on the user-defined separate symbol (e.g., <sep>),
and then, represent each sentence as a set of word embedding vectors, $W_j=\{w_1, w_2, ..., w_n\}$, 
where $W_j$ is the set of word embedding vectors in the $j^{th}$ sentence, $w_i$ is the $d$-dimensional embedding vector of the $i^{th}$ word, 
and $n$ is the maximum number of words in a single sentence.
Once all news articles in the input dataset are pre-processed, 
we initialize all word embedding vectors, $\hat{W}=\{w_1, w_2, ..., w_N\}$, where $N$ is the total number of words appearing in the entire dataset.

\vspace{1mm}
\noindent
\textbf{Word-level attention layer.}
In the word-level attention layer, 
we aim to learn `the relationships among words' in the same sentence, instead of in the entire article, to capture the \textit{local} context of each sentence.
Specifically, 
we apply multi-head self-attention blocks to the set of word embedding vectors in each sentence (i.e., sentence-wise self-attention).
Thus, 
the word-level attention layer for the $j^{th}$ sentence is defined as follows:
\begin{equation}
\Tilde{W}_j = MultiHead(Q^w, K^w, V^w),
\label{eq:word-level}
\end{equation}
where $Q^w=K^w=V^w=W_j$ and $W_j$ is the set of word embedding vectors in the $j^{th}$ sentence.
Then, we pass the output ($\Tilde{W}_j$) -- the \textit{local-context-aware} word embedding vectors in the $j^{th}$ sentence --
to a feed-forward layer, with the same principle of~\cite{vaswani2017attention}, 
in order to represent the output from the word-level attention layer to better fit the input for the next sentence-level attention layer.

Note that this `sentence-wise' attention layer of {\han} has advantages in terms of both efficiency and effectiveness, 
compared to the existing `article-wise' attention methods~\cite{du2017stance,chen2022mulan} that take into account all words in the article together.
{\han} (1) requires the amount of computation much less than the existing methods (\textit{efficient}) 
and (2) is able to effectively capture the local context without being interfered with the words that are located far away (\textit{effectiveness})
because {\han} only learns the relationships among closely located words.
However, it is difficult to capture the \textit{global} context of a news article with only the word-level attention layer because this layer is designed specially for capturing the local context of each sentence.
To address this limitation, we apply the sentence-level attention layer to the output of this word-level attention layer.

\vspace{2mm}
\noindent
\textbf{Sentence-level attention layer.}
The sentence-level attention layer of {\han} learns `the relationships among sentences' in the entire article, 
to capture the \textit{global} context of the news article.
First, the word embedding vectors from the previous layer are averaged in a sentence-wise manner to generate the sentence embedding vectors.
Thus, the $j^{th}$ sentence embedding vector ($s_j$) with $d$-dimensionality is generated by averaging the $d$-dimensional word embedding vectors in the $j^{th}$ sentence (i.e., $s_j = Avg(\Tilde{W}_j)$),
where each sentence embedding vector ($s_j$) has the local context information captured from the previous layer.
Then, the sentence-level multi-head self-attention blocks are applied to the set of sentence embedding vectors in each article (i.e., article-wise self-attention).
Formally, 
the sentence-level attention layer for the $k^{th}$ article is defined as the similar way to the word-level attention:
\begin{equation}
\Tilde{S}_k = MultiHead(Q^s, K^s, V^s),
\label{eq:sentence-level}
\end{equation}
where $Q^s=K^s=V^s=S_k$ and $S_k$ is the set of the sentence embedding vectors in the $k^{th}$ article.
We also pass the \textit{global-context-aware} sentence embedding vectors in $k^{th}$ article ($\Tilde{S}_k$) to a feed-forward layer to process the output from this sentence-level attention layer to better fit the next layer (i.e., the title-attention layer).

For the learnable parameters of the word-level and sentence-level attention layers,
we use the different sets of parameters in the word-level and the sentence-level attention layers (i.e., $X^{Q^w} \neq X^{Q^s}$)
because the related patterns among words (\textit{the local context}) to capture in the word-level layer are highly likely to be different from those among sentences (\textit{the global context}) to capture in the sentence-level layer.
As a result, this sentence-level attention layer is able to effectively capture the global context of a news article,
based on the previously learned local context, 
compared to existing non-hierarchical methods counting on all words at once.

\vspace{2mm}
\noindent
\textbf{Title-level attention layer.}
The title of a news article, a special sentence having the key idea that the author of the news article hopes to deliver, 
has been considered as important information in a number of news recommendation systems~\cite{wang2018dkn,ge2020graph,zhu2019dan,wu2019npa,zhang2021combining,kim2022title}.
Inspired by the importance of the title, 
we apply the title-level attention layer to the sentence embeddings.
The goal of this title-attention layer is (1) to reinforce the key idea included in the title and (2) to filter out relatively unnecessary information in sentences.
Let $T_k$ be the title embedding vector of the $k^{th}$ article,
then, the title-level attention layer for the $k^{th}$ article is defined as follows:
\begin{equation}
\Tilde{S}^T_k = MultiHead(Q^t, K^t, V^t),
\label{eq:title-level}
\end{equation}
where $Q^t=T_k$, $K^t=V^t=\Tilde{S}_k$, and $\Tilde{S}_k$ is the sentence embedding vectors from the previous layer.
Thus, by Eq~\ref{eq:title-level}, the sentence embedding vectors would be re-weighted in terms of the context of the title (i.e., reinforcing the key idea in the title).
Applying the title attention layer in the final stage of {\han}, however, 
might cause the global-context-aware sentence embedding vectors to be biased only to the context of the title too much,
thereby leading to the degradation of prediction accuracy.
To prevent this problem, 
we add a residual connection~\cite{he2016deep,wei2018residual} to the output of the title attention layer in order to maintain the global context previously learned from the sentence-level attention layer.
Thus, the final output of the title-level attention layer is defined as follows:
\begin{equation}
\Tilde{S}^*_k = \Tilde{S}^T_k + \Tilde{S}_k,
\label{eq:residual-title}
\end{equation}
where $\Tilde{S}^*_k$ is the set of the final sentence embeddings in the $k^{th}$ article.
Finally, we aggregate the sentence embeddings, and then pass the aggregated embedding through the output layer.
\begin{equation}
\hat{y} = Predict(a_k), \hspace{2mm} a_k = Avg(\Tilde{S}^*_k),
\label{eq:article-average}
\end{equation}
where $Predict(\cdot)$ is a softmax layer to predict the political stance of a given news article, $\hat{y}$.

As a result, via the 3-level hierarchical process,
{\han} is able to effectively capture both the local and global context implicitly reflected in a news article.
We will verify the effectiveness of {\han} in improving the model accuracy of {\m} in Section~\ref{sec:eval}.

\subsection{Knowledge Encoding ({\ke})}\label{sec:proposed-ke}
As shown in Appendix~\ref{sec:appendix-userstudy}, 
it is crucial to understand the \textit{interpretation} and \textit{sentiment} to real-world entities such as keywords and persons in predicting the political stance of a news article.
In general, however, it often occurs that the information about many real-world entities is not clearly provided in news articles.
For example, famous politicians (e.g., `Barack Obama' and `Donald Trump') are not always directly described in a news article.
From this motivation, 
we propose \textbf{\underline{K}}nowledge \textbf{\underline{E}}ncoding (\textbf{{\ke}}) that pre-trains the external knowledge (both common and political) related to the real-world entities and injects the external knowledge into the corresponding words appearing in a news article for accurate political stance prediction.
Regarding political knowledge, 
we take into account two different political knowledge (i.e., liberal and conservative) separately in the knowledge encoding.

This approach has advantages in capturing subtle but important difference in knowledge, related to real-world entities, varying depending on political stances,
compared to the existing knowledge-based methods~\cite{zhang2022kcd,feng2021kgap} that consider only unified political knowledge.
However, there is a technical challenge about how to incorporate the three different knowledge (one common and two political knowledge) into the process of the political stance prediction.
To this challenge, we propose a simple but effective algorithm for knowledge injection, 
where {\ke} learns how to fuse the three types of knowledge.
Figures~\ref{fig:overview}(e)-(g) illustrate the three stages of {\ke}:
(1) knowledge preparation; (2) embedding; and (3) injection. 

\begin{table}[t]
\centering
\caption{Statistics of political KGs.}
\vspace{-3mm}
\label{table:kg}
\setlength\tabcolsep{5pt}
\begin{tabular}{c||c|cc}
\toprule
 
 & KGAP~\cite{feng2021kgap} & \textsf{KG-lib} & \textsf{KG-con} \\

\midrule
\# of source posts  & -  & 219,915 & 276,156 \\
\# of entities  & 1,071 & 5,581 & 6,316 \\
\# of relations  &  10,703 & 29,967 & 33,207 \\
Political stances  & Both & Liberal & Conservative \\

\bottomrule
\end{tabular}
\end{table}

\vspace{1mm}
\noindent
\textbf{Knowledge preparation.}
First, we prepare both common and political knowledge for real-world entities appearing in news articles.
In terms of common knowledge,
there are many well-designed knowledge graphs (KG) such as YAGO~\cite{pellissier2020yago}, Freebase~\cite{bollacker2008freebase}, and WordNet~\cite{miller1995wordnet} built from large-scale real-world datasets, 
where a node represents an entity and an edge represents a relation between two entities.
Among them, we choose YAGO~\cite{pellissier2020yago} as the source of common knowledge in this work
because YAGO consists of general knowledge about real-world entities (e.g., people, cities, countries, movies, and organizations).
On the other hand, political knowledge has been rarely studied except for some works~\cite{feng2021kgap,zhang2022kcd} that built a single political KG based on the political entities and their relations from U.S. political websites (e.g., AFL-CIO and Heritage Action)
Unfortunately, it is challenging to accurately represent the relations among political entities in a single knowledge graph
because \textit{the interpretation and sentiment to political entities and their relations can be different depending on the political stance}.
To address this challenge, 
we (1) collected 496,071 political-related posts from the U.S. political community websites\footnote{\url{https://www.reddit.com/r/Liberal/}, \url{https://www.reddit.com/r/Conservative/}},
(2) extracted 18 political entities, using an NER (named entity recognition) method~\cite{vychegzhanin2019comparison},
and (3) constructed two different political knowledge graphs: \textsf{KG-lib} and \textsf{KG-con}, 
where each data point is represented as a triplet: <head entity, relation, tail entity>.
Table~\ref{table:kg} shows the statistics of the political knowledge graphs.
The details about the knowledge graph construction are included in Appendix~\ref{sec:appendix-data}.

\noindent
\textbf{Knowledge embedding.}
Next, we apply knowledge embedding methods~\cite{bordes2013translating,dettmers2018convolutional,sun2018rotate,zhang2020learning,lee2022thor}, aiming to learn the relations among entities, to the three KGs independently, 
in order to represent three types of embedding vectors for each entity: $K^{com}$, $K^{lib}$, and $K^{con}$  (i.e., common, liberal, and conservative).
Since this work is \textit{agnostic} to the knowledge embedding method, 
any knowledge embedding methods could be applied to {\ke}.
As the knowledge embedding method, we consider three recent methods: RotatE~\cite{sun2018rotate}, ModE~\cite{zhang2020learning}, and HAKE~\cite{zhang2020learning}.
In this paper, we omit the details of the knowledge embedding methods because it is beyond the scope of this work, but we include the knowledge graph completion accuracies of the three knowledge embedding methods in Appendix~\ref{sec:appendix-additional-exp}.

\vspace{1mm}
\noindent
\textbf{Knowledge injection.}
As we mentioned in the beginning of Section~\ref{sec:proposed-ke},
there is a technical challenge about how to incorporate the three different knowledge into the process of predicting the political stance of a news article.
Specifically, it is difficult to determine (1) how to fuse the three types of knowledge and (2) how much amount of each knowledge is needed, for accurate political stance prediction.
To this challenge, 
we propose a simple but effective knowledge injection algorithm that fuses the three types of knowledge for real-world entities and injects them into the corresponding words appearing in a news article.

\begin{algorithm}[t]
\caption{Knowledge injection of {\ke}}
\begin{algorithmic}[1]
\REQUIRE news article $a$, word embeddings $W$, three types of knowledge embeddings: $K^{com}$, $K^{lib}$, $K^{con}$, knowledge control factors $\alpha, \beta$
\STATE Initialize $W^* \leftarrow \phi$ 
\STATE \textbf{for} word $i \in a$ \textbf{do}
\STATE\hspace{1.5em} $e \leftarrow W[i]$
\STATE\hspace{1.5em} $e^{com} \leftarrow (1-\alpha)\cdot e \oplus \alpha\cdot K^{com}[i]$
\STATE\hspace{1.5em} $e^{lib} \leftarrow (1-\beta)\cdot e^{com} \oplus \beta\cdot K^{lib}[i]$
\STATE\hspace{1.5em} $e^{con} \leftarrow (1-\beta)\cdot e^{com} \oplus \beta\cdot K^{con}[i]$

\STATE\hspace{1.5em} $W^*[i] \leftarrow \textsf{Fuse}([e^{lib} \| e^{con}]) \oplus e$
\STATE \textbf{end for}    
\STATE \algorithmicreturn \hspace{0.5mm} $W^*$
\end{algorithmic}
\label{algo:knowledge-injection}
\end{algorithm}

The process of the knowledge injection of {\ke} is illustrated in Figure~\ref{fig:overview}(g) and described in Algorithm~\ref{algo:knowledge-injection}.
For each word in a given article,
{\ke} first injects the common knowledge of the word ($K^{com}[i]$) into the corresponding word embedding (lines 3-4 in Algorithm~\ref{algo:knowledge-injection}),
where $\alpha$ is the common knowledge control factor and `$\oplus$' means the element-wise addition.
Then, we add the political knowledge embeddings to the common knowledge injected embedding $e^{com}$ with the political knowledge control factor $\beta$ independently (lines 5-6).
The two embeddings with different political knowledge, $e^{lib}$ and $e^{con}$, are concatenated and passed through a fully-connected layer, \textsf{Fuse}$(\cdot)$ that fuses the two embeddings into a single knowledge-aware embedding (line 7),
which plays a role to learn how to fuse the different political knowledge for accurate prediction.
We also add the original word embedding to the knowledge-aware embedding (i.e., residual connection).
As a result, the knowledge-aware word embeddings can be used across the hierarchical news encoding process of {\han}.
We will verify the effectiveness of {\ke} and the impacts of its hyperparameters $\alpha$ and $\beta$ in improving the political stance prediction accuracy of {\m} in Section~\ref{sec:eval}.

\section{Experimental Validation}\label{sec:eval}
In this section, we comprehensively evaluate {\m} by answering the following evaluation questions: 

\begin{itemize}[leftmargin=10pt]
    \item EQ1. To what extent does {\m} improve existing methods in terms of the model accuracy in the political stance prediction?
    \item EQ2. How fast does {\m} converge in terms of time and epochs?
    \item EQ3. How effective are the main components of {\m} in terms of improving the model accuracy in political stance prediction?
    \item EQ4. How sensitive is the model accuracy of {\m} to the hyperparameters $\alpha$ and $\beta$?
\end{itemize}

\subsection{Experimental Setup}\label{sec:eval-setup}
\noindent
\textbf{Datasets.} 
We evaluate {\m} with three real-world news article datasets, \textsf{SemEval}~\cite{kiesel2019semeval}, \textsf{AllSides-S}~\cite{li2019encoding}, and \textsf{AllSides-L}.
Table~\ref{table:datasets} shows the statistics of the news article datasets.
\textsf{SemEval} consists of 645 articles with 2 classes (hyperpartisan and center) and \textsf{AllSides-S} consists of 14,783 articles with 3 classes (left, center, and right).
For training, we use 10-fold and 3-fold cross validations for \textsf{SemEval} and \textsf{AllSides-S}, respectively, as the same in previous works~\cite{li2019encoding,li2021using,feng2021kgap,zhang2022kcd}.
For extensive evaluation, we construct a large-scale political news dataset, 
\textsf{AllSides-L} with 719,256 articles with 5 classes (left, lean left, center, lean right, and right)\footnote{The data construction details for \textsf{AllSides-L} are provided in Appendix~\ref{sec:appendix-data}.}.
We split the \textsf{AllSides-L} dataset into training and validation sets: 
647,330 articles in the training set and 71,926 articles in the validation set.

\vspace{0.05in}
\noindent
\textbf{Baseline methods.}
We compare {\m} with seven baseline methods: five text-based methods~\cite{mikolov2013efficient,pennington2014glove,peters-etal-2018-deep,kenton2019bert,liu2019roberta} and two knowledge-based political stance prediction methods~\cite{feng2021kgap,zhang2022kcd}.
Word2Vec~\cite{mikolov2013efficient}, GloVe~\cite{pennington2014glove}, and ELMo~\cite{peters-etal-2018-deep} are general language models that aim to capture context from text.
We also use pre-trained BERT~\cite{kenton2019bert} and RoBERTa~\cite{liu2019roberta} by fine-tuning on training datasets.
KGAP~\cite{feng2021kgap} is a knowledge-aware approach that leverages a political knowledge graph with graph neural networks (e.g., R-GCN~\cite{schlichtkrull2018modeling}).
KCD~\cite{zhang2022kcd}, the state-of-the-art model, considers knowledge walks from a political knowledge graph and textual cues in political stance prediction.

\begin{table}[t]
\centering
\caption{Statistics of political news article datasets.}
\vspace{-3mm}
\label{table:datasets}
\setlength\tabcolsep{3pt}
\begin{tabular}{c||c|c}
\toprule
 
Dataset & \# of articles & Class distribution  \\

\midrule
\textsf{SemEval} & 645 & 407 / 238 \\
\textsf{AllSides-S} & 14.7k & 6.6k / 4.6k / 3.5k \\
\textsf{AllSides-L} & 719.2k & 112.4k / 202.9k / 99.6k / 62.6k / 241.5k \\

\bottomrule
\end{tabular}
\end{table}

\vspace{0.05in}
\noindent
\textbf{Implementation details.} 
We use PyTorch 1.10.0 to implement all methods including {\m} on Ubuntu 20.04 OS.
We run our experiments on the machine equipped with an Intel i7-9700k CPU with 64 GB memory and a NVIDIA RTX 2080 Ti GPU, installed with CUDA 11.3 and cuDNN 8.2.1.
We set the batch size $b$ as 16 for all datasets as the same in the previous works~\cite{feng2021kgap,zhang2022kcd}.
We use the Adam optimizer~\cite{kingma2015adam} with the learning rate $\eta=\text{1e-3}$ and the weight decay factor 5e-2 for all datasets.
As a learning rate scheduler~\cite{ko2022lena,goyal2017accurate,johnson2020adascale},
we use the `ReduceLROnPlateau' scheduler of PyTorch that reduces the learning rate $\eta$ by 1/2
when the training loss has stopped improving for a `patience' number of epochs in a row (we set the patience epoch as 5).

\begin{table}[t!]
\centering
\caption{Comparison of the model accuracy on three real-world datasets (The bold font indicates the best results).}
\vspace{-3mm}
\label{table:model-accuracy}
\setlength\tabcolsep{8pt}
\begin{tabular}{c||c|c|c}
\toprule
\multirow{2}{*}{Method} & \multicolumn{3}{c}{Dataset} \\ 
\cmidrule(lr){2-4} 
 & \textbf{SemEval} & \textbf{AllSides-S}  & \textbf{AllSides-L} \\ 

\midrule

\textbf{Word2Vec}~\cite{mikolov2013efficient}    & 0.7027 & 0.4858 & 0.4851  \\
\textbf{GloVe}~\cite{pennington2014glove}       & 0.8071 & 0.7101 & 0.6354  \\
\textbf{ELMo}~\cite{peters-etal-2018-deep}        & 0.8678 & 0.8197 & 0.7483  \\
\textbf{BERT}~\cite{kenton2019bert}        & 0.8692 & 0.8246 & 0.7812  \\
\textbf{RoBERTa}~\cite{liu2019roberta}     & 0.8708 & 0.8535 & 0.8222  \\

\midrule

\textbf{KGAP}~\cite{feng2021kgap}    & 0.8956 & 0.8602 & N/A \\
\textbf{KCD}~\cite{zhang2022kcd}     & 0.9087 & 0.8738 & N/A \\

\midrule

\textbf{{\m}}-RotatE & 0.9426 & 0.9151 & 0.8584 \\
\textbf{{\m}}-HAKE & 0.9395 & 0.9216 & 0.8563 \\
\textbf{{\m}}-ModE & \textbf{0.9521} & \textbf{0.9256} & \textbf{0.8617} \\

\bottomrule
\end{tabular}
\end{table}


\subsection{Experimental Results}\label{sec:eval-result}

\textbf{EQs 1-2. Accuracy and Efficiency}.
In this experiment, we evaluate the accuracy and efficiency of the proposed {\m} in the political stance prediction.
We train {\m} on the three real-world political news article datasets (50 epochs) with varying knowledge embedding methods (RotatE, HAKE, and ModE).
For the \textsf{SemEval} and \textsf{AllSides-L} datasets,
we apply \textit{k}-fold cross validations (10-fold for \textsf{SemEval} and 3-fold for \textsf{AllSides-S}).
Due to the page limit, we report the averaged accuracy of \textit{k}-fold cross validations and include the entire results with the standard deviation in Appendix~\ref{sec:appendix-additional-exp}.
For \textsf{AllSides-L},
we train {\m} on the training set (647k articles) and measure the model accuracy using the validation set (71.9k articles), which is not used in the training process.

As shown in Table~\ref{table:model-accuracy}\footnote{For \textsf{SemEval} and \textsf{AllSides-S}, we use the results for baseline methods reported in~\cite{zhang2022kcd}. 
We also obtain the results for baseline methods and include them in Appendix~\ref{sec:appendix-additional-exp}.},
{\m} consistently outperforms all baseline methods in terms of the model accuracy, regardless of knowledge embedding methods.
Specifically, {\m} improves the state-of-the-art method, KCD~\cite{zhang2022kcd} by 4.77\% and 5.92\% in \textsf{SemEval} and \textsf{AllSides-S} datasets, respectively.
These improvements over KCD are significant,
given that KCD has already achieved quite high accuracies in those datasets. 
We also evaluate {\m} on a large-scale dataset (\textsf{AllSides-L}) which is $48\times$ larger and has more classes (i.e., more difficult to predict) than the previous largest one (\textsf{AllSides-S}).
{\m} still significantly outperforms all baseline methods although the accuracy of {\m} decreases to some extent in \textsf{AllSides-L}, compared with the other datasets.
In addition, we have conducted the \textit{t}-tests with a 95\% confidence level
and verified that the improvement of {\m} over all baseline methods are statistically significant (i.e., the \textit{p}-values are below 0.05).
As a result, {\m} consistently outperforms state-of-the-art methods in all datasets, including a our own dataset (\textsf{AllSides-L}), with a 95\% confidence level,
which implies that {\m} has a good generalizability.
In addition to generalizability, in terms of applicability,
{\m} could be applied to other languages since {\m} does not depend on any linguistic features (e.g., word-order and grammatical features). 
As an example, we have successfully applied {\m} to a web-based platform for diverse political news consumption in non-English.

\begin{figure}[t]
\begin{subfigure}[b]{0.3\textwidth}
    \includegraphics[width=\linewidth]{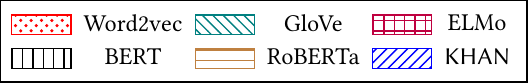}
\end{subfigure}

\begin{subfigure}[b]{0.25\textwidth}
    \includegraphics[width=0.9\linewidth]{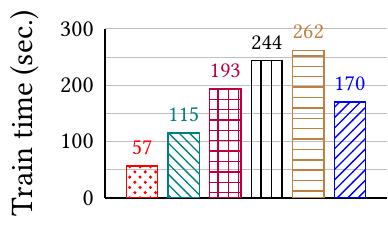}
    \vspace{-3mm}
    \caption{\textsf{SemEval}}
\end{subfigure}%
\begin{subfigure}[b]{0.25\textwidth}
    \includegraphics[width=0.9\linewidth]{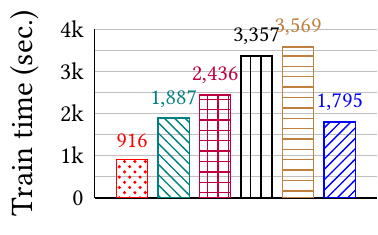}
    \vspace{-3mm}
    \caption{\textsf{AllSides-S}}
\end{subfigure}
\begin{subfigure}[b]{0.25\textwidth}
    \includegraphics[width=0.9\linewidth]{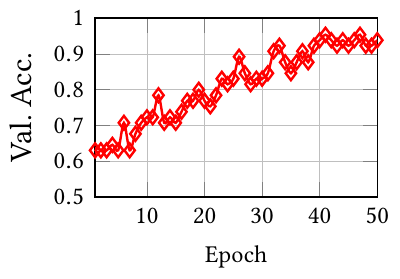}
    \vspace{-3mm}
    \caption{\textsf{SemEval}}
\end{subfigure}%
\begin{subfigure}[b]{0.25\textwidth}
    \includegraphics[width=0.9\linewidth]{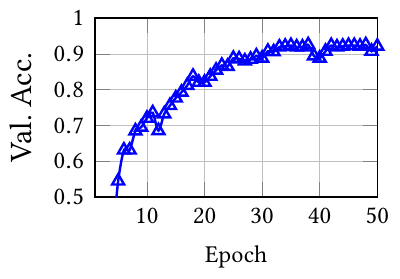}
    \vspace{-3mm}
    \caption{\textsf{AllSides-S}}
\end{subfigure}%
\vspace{-3mm}
\caption{The training time and convergence rate with respect to training epochs on SemEval and AllSides-S.}\label{fig:eval-performance}
\vspace{-5mm}
\end{figure}

We also evaluate {\m} in terms of the training time and convergence rate with respect to training epoch.
Figures~\ref{fig:eval-performance}(a)-(b) represent the training time for 50 epochs of each method and Figures~\ref{fig:eval-performance}(c)-(d) show the validation accuracy of {\m} with respect to the training epoch.
{\m} finishes its training in shorter time than ELMo, BERT, and RoBERTa (even faster than GloVe in \textsf{AllSides-S}),
while achieving the highest accuracy at the same time.
Considering the low accuracies of Word2Vec and GloVe (at least 15\% less than {\m}), these results verify the efficiency of {\m}.
In terms of convergence rate, {\m} converges to high accuracy within only around 40 epochs (Figure~\ref{fig:eval-performance}(c)-(d)).
All in all, these results demonstrate 
that {\m} is able to \textit{effectively} and \textit{efficiently} solve the political stance prediction problem by employing the proposed {\han} and {\ke}.

\begin{figure*}[t]
\centering
\begin{subfigure}[b]{0.5\textwidth}
    \includegraphics[width=\linewidth]{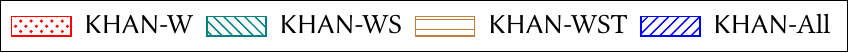}
\end{subfigure}
\begin{subfigure}[b]{0.33\textwidth}
    \includegraphics[width=\linewidth]{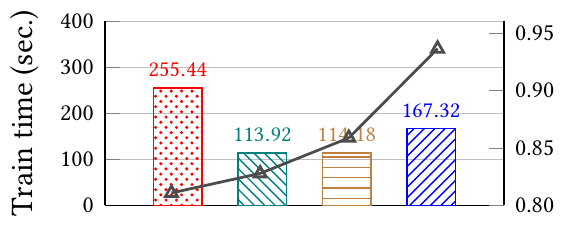}
    \vspace{-8mm}
    \caption{\textsf{SemEval} ($d=128$)}
\end{subfigure}%
\begin{subfigure}[b]{0.29\textwidth}
    \includegraphics[width=\linewidth]{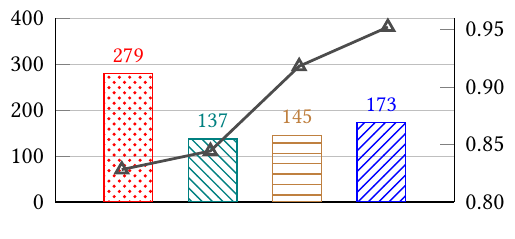}
    \vspace{-8mm}
    \caption{\textsf{SemEval} ($d=256$)}
\end{subfigure}%
\begin{subfigure}[b]{0.33\textwidth}
    \includegraphics[width=\linewidth]{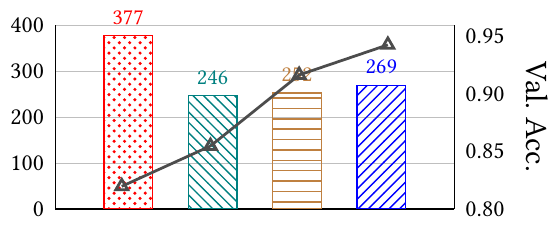}
    \vspace{-8mm}
    \caption{\textsf{SemEval} ($d=512$)}
\end{subfigure}
\begin{subfigure}[b]{0.33\textwidth}
    \includegraphics[width=\linewidth]{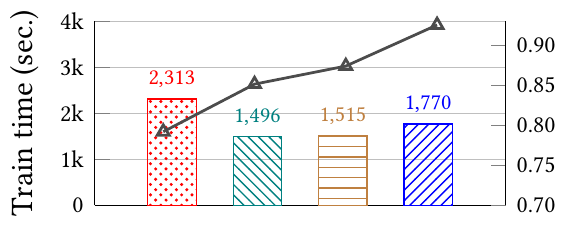}
    \vspace{-8mm}
    \caption{\textsf{AllSides-S} ($d=128$)}
\end{subfigure}%
\begin{subfigure}[b]{0.29\textwidth}
    \includegraphics[width=\linewidth]{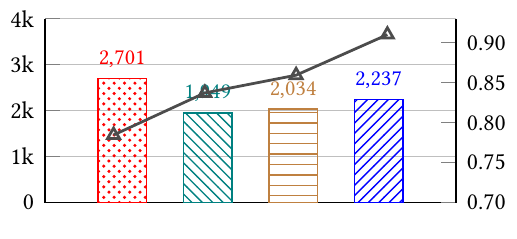}
    \vspace{-8mm}
    \caption{\textsf{AllSides-S} ($d=256$)}
\end{subfigure}%
\begin{subfigure}[b]{0.33\textwidth}
    \includegraphics[width=\linewidth]{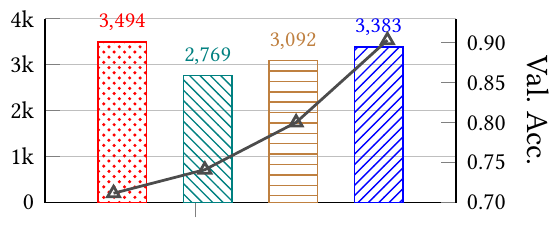}
    \vspace{-8mm}
    \caption{\textsf{AllSides-S} ($d=512$)}
\end{subfigure}%
\vspace{-3mm}
\caption{Effectiveness of the main components of {\m} in terms of the training time (bar) and model accuracy (line).}\label{fig:eval-ablation}
\end{figure*}

\begin{figure*}
    \centering
    \begin{subfigure}[c]{0.248\textwidth}
        \includegraphics[width=\linewidth]{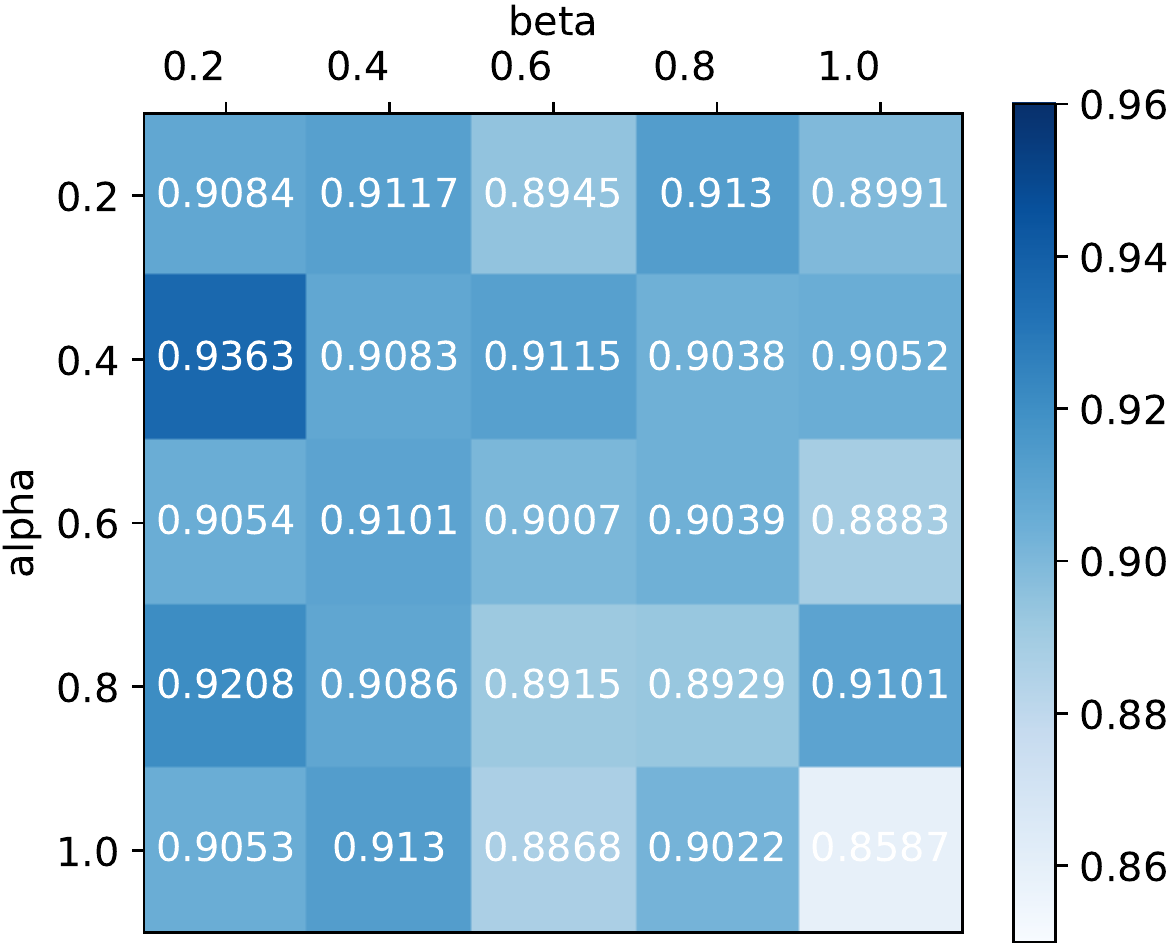}
        \caption{\textsf{SemEval} ($d=128$)}
    \end{subfigure}%
    \begin{subfigure}[c]{0.248\textwidth}
        \includegraphics[width=\linewidth]{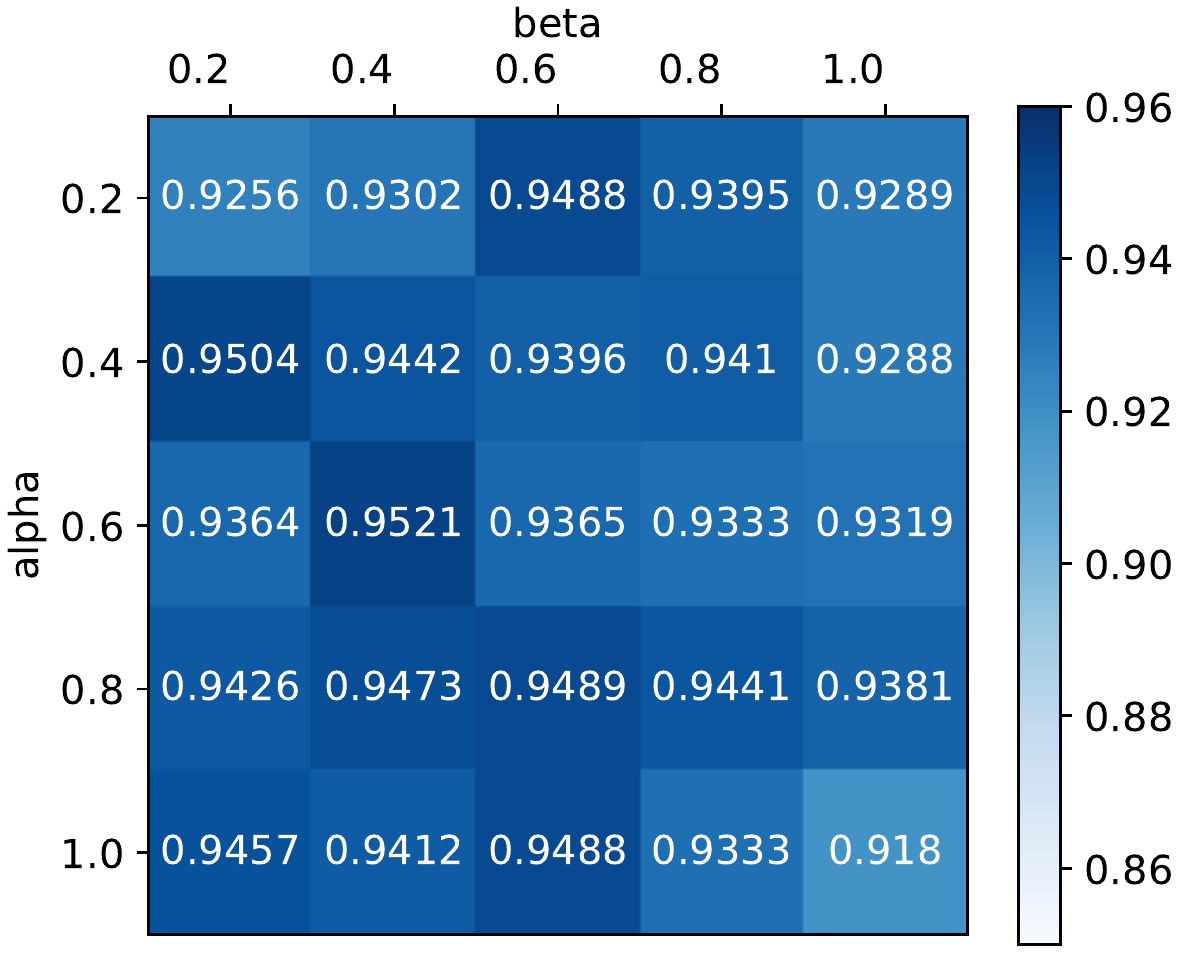}
        \caption{\textsf{SemEval} ($d=256$)}
    \end{subfigure}%
    \begin{subfigure}[c]{0.248\textwidth}
        \includegraphics[width=\linewidth]{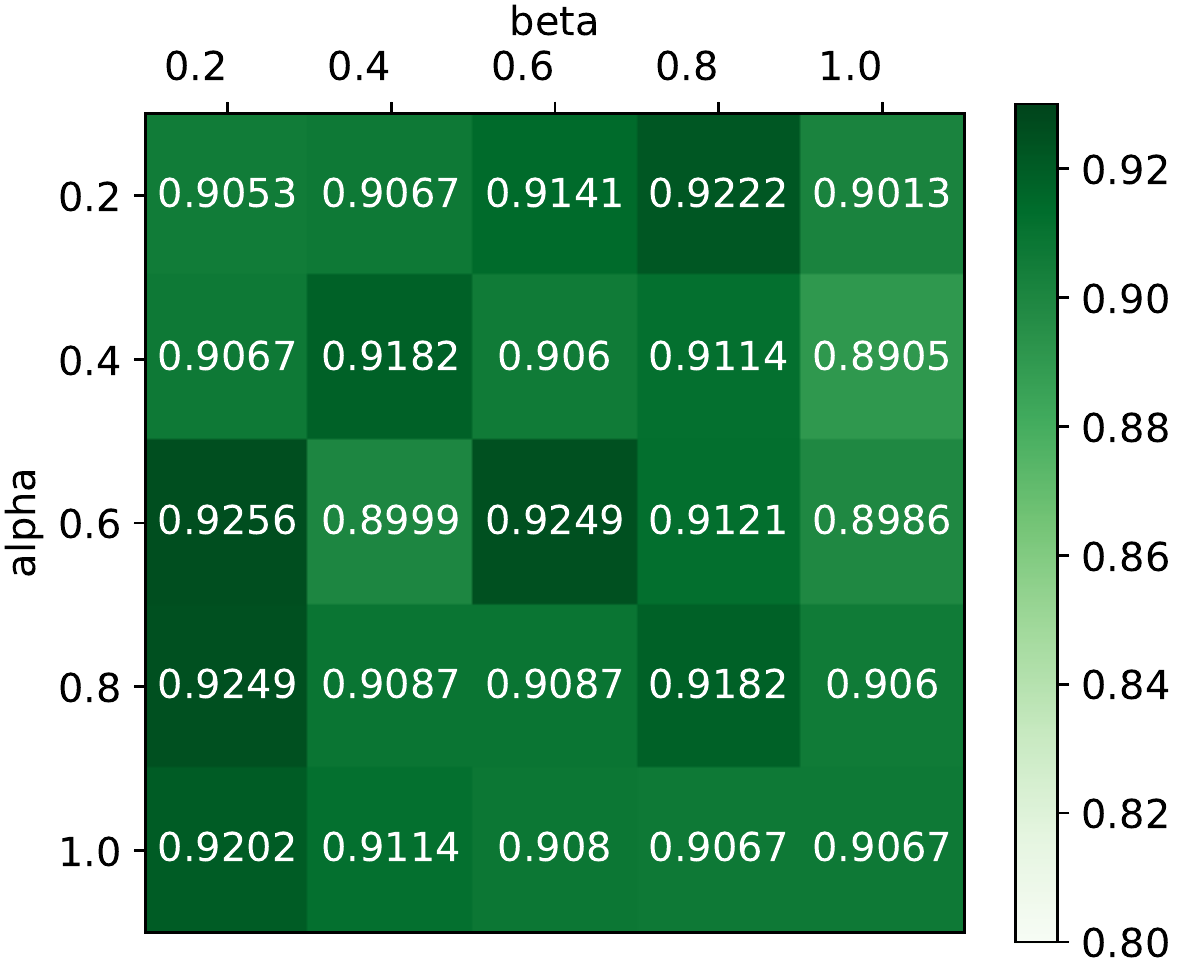}
        \caption{\textsf{AllSides-S} ($d=128$)}
    \end{subfigure}%
    \begin{subfigure}[c]{0.248\textwidth}
        \includegraphics[width=\linewidth]{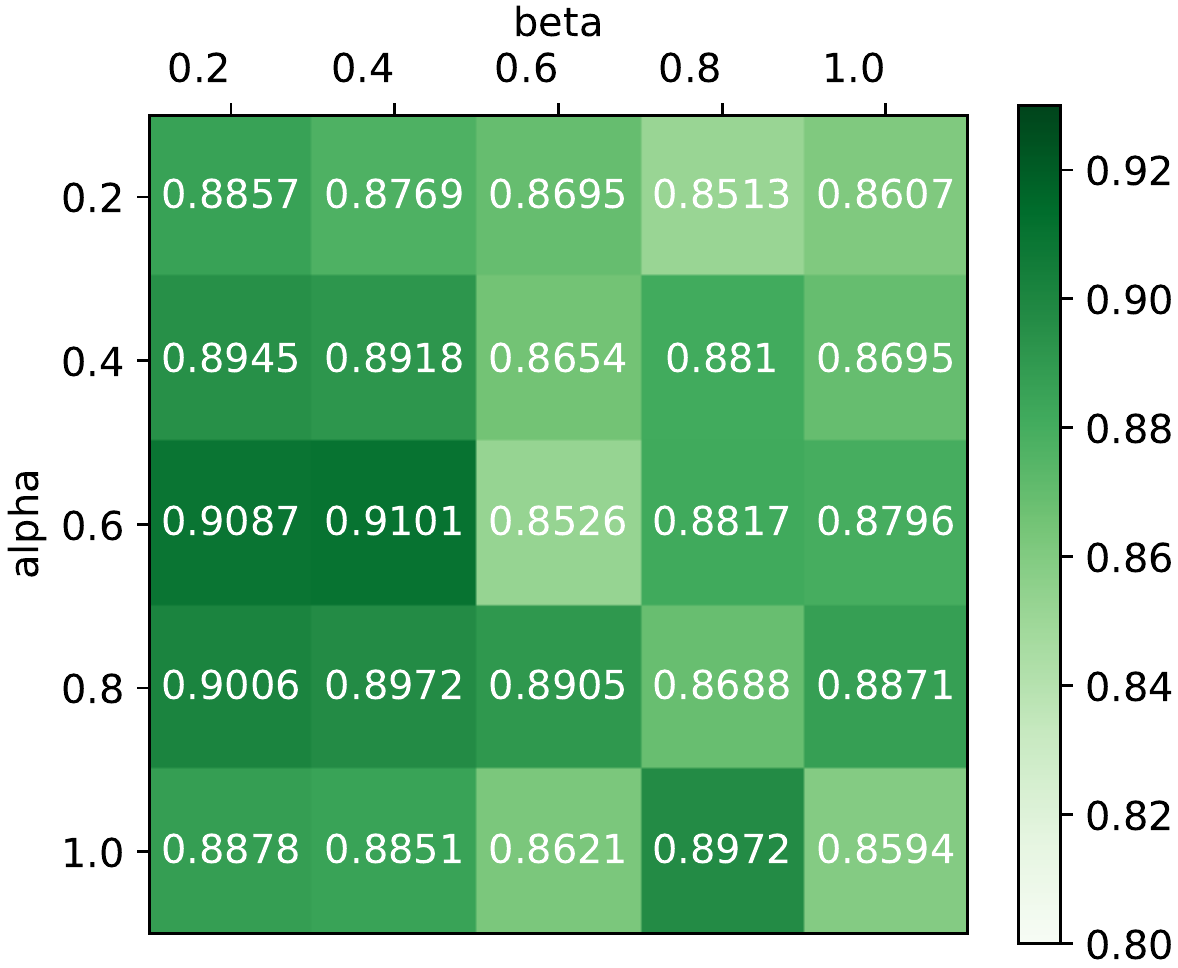}
        \caption{\textsf{AllSides-S} ($d=256$)}
    \end{subfigure}%
    \vspace{-3mm}
    \caption{The impact of hyperparameters $\alpha$ and $\beta$ on the model accuracy of {\m} in political stance prediction.}
    \vspace{-3mm}
    \label{fig:hyperparameter}
\end{figure*}

\vspace{2mm}
\noindent
\textbf{EQ3. Ablation Study}.
In this experiment, we verify the effectiveness of the {\han} and {\ke}. 
We compare the four versions of {\m}:

\begin{itemize}[leftmargin=12pt]
    \item {\m}-W: a baseline with only the word-level attention layer.
    \item {\m}-WS: the version with the word and sentence layers.
    \item {\m}-WST: the version with {\han} (word, sentence, and title).
    \item {\m}-All: the original version with both {\han} and {\ke}.
\end{itemize}
We train each version of {\m} with varying the embedding dimensionality ($d=128$, $256$, and $512$) on \textsf{SemEval} (10-fold) and \textsf{AllSides-S} (3-fold), and measure the accuracy and training time.
As shown in Figure~\ref{fig:eval-ablation},
{\m}-WS consistently improves {\m}-W in both the accuracy and training performance.
This result demonstrates that our hierarchical approach not only (1) (\textit{efficiency}) requires much less computation overhead than non-hierarchical existing methods, 
but also (2) (\textit{effectiveness}) captures the context implicitly reflected in a news article successfully, 
as we claimed in Section~\ref{sec:proposed-han}.
{\m}-WST further improves the model accuracy with only a very small amount of overhead,
thereby verifying the importance of the title that has the key idea of the news article in political stance prediction.
Finally, 
{\m}-All always achieves the highest accuracy in all cases, regardless of datasets and embedding dimensionality.
This result validates that external knowledge for real-world entities appearing in a news article is indeed important in predicting its political stance, as we claimed in Section~\ref{sec:proposed-ke}.

\noindent
\textbf{EQ4. Hyperparameter sensitivity}.
Finally, we evaluate the impacts of the hyperparameters $\alpha$ and $\beta$ on the model accuracy of {\m}.
As explained in Section~\ref{sec:proposed-ke}, 
the hyperparameter $\alpha$ ($\beta$) controls the amount of common (political) knowledge is injected to the corresponding words in a given news article.
Thus, as $\alpha$ ($\beta$) becomes lower, the more amount of the common (political) knowledge is injected to the entities appearing in a new article. 
While, as $\alpha$ ($\beta$) becomes larger, the less amount of the common (political) knowledge is injected. 
For an extreme case, if $\alpha$ ($\beta$) is 1, the common (political) knowledge is not used.
We measure the model accuracy with varying $\alpha$ and $\beta$ from 0.2 to 1.0 on \textsf{SemEval} and \textsf{AllSides-S}.

As clearly illustrated in Figure~\ref{fig:hyperparameter},
{\m} with lower $\beta$ tends to achieve higher accuracy than {\m} with larger $\beta$ (i.e., the left-hand side of each heatmap tends to be darker than its right-hand side in each row).
On the other hand, {\m} with larger $\alpha$ and $\beta$ shows relatively poor performance in political stance prediction accuracy (i.e, the bottom right side of each heatmap tends to be brighter than other sides). 
Specifically, {\m} with $\alpha=1$ and $\beta=1$ (i.e., both common and political knowledge not used) shows the worst results in the \textsf{SemEval} dataset.
These results verify that external knowledge for real-world entities is indeed useful in predicting the political stance of a news article.
Note that {\m} with $\alpha$ and $\beta$ below $0.6$ consistently outperforms all baseline methods.
Based on these results, 
we believe that the model accuracy of {\m} is not sensitive to the hyperparameters $\alpha$ and $\beta$, and we recommend to set the hyperparameters $\alpha$ and $\beta$ as below $0.6$.

\section{Conclusion}\label{sec:conclusion}
In this paper, via a carefully-designed user study,
we observe that both explicit and implicit factors (i.e., the context/tone and external knowledge for real-world entities) are important in predicting the political stances of news articles.
Based on the observations, 
we propose a novel approach to accurate political stance prediction, {\m} that successfully captures the local and global context of a new article with the two key components:
(1) hierarchical attention networks ({\han}) to learn the relationships among words, sentences, and the title in a news article with the 3-level hierarchy 
and (2) knowledge encoding ({\ke}) to incorporate the three types of useful knowledge for real-world entities into the process of the political stance prediction.
Via the extensive experiments,
we demonstrate that 
(1) (\textit{accuracy}) {\m} consistently achieves higher accuracies than all baseline methods,
(2) (\textit{efficiency}) {\m} is able to converge to high accuracies within comparable training time (epochs),
and (3) (\textit{effectiveness}) the key components of {\m} ({\han} and {\ke}) are quite effective in improving the model accuracy of {\m}.

\begin{acks}
This work was supported by the National Research Foundation of Korea (NRF) (2018R1A5A7059549), and Institute of Information \& Communications Technology Planning \& Evaluation (IITP) (RS-2022-00155586, 2022-0-00352, 2020-0-01373).
Hanghang Tong was partially supported by NSF (1947135, 2134079, and 1939725).
\end{acks}


\balance

\bibliographystyle{ACM-Reference-Format}
\bibliography{bibliography}

\clearpage
\appendix
\section{Appendix}\label{sec:appendix}

In this appendix, we describe the detailed information of our user study (Appendix~\ref{sec:appendix-userstudy}), data construction (Appendix~\ref{sec:appendix-data}), and the additional experimental results about the reliability of our experiments about political stance prediction (Appendix~\ref{sec:appendix-additional-exp}).

\subsection{User Study}\label{sec:appendix-userstudy}
In this section, we describe the details of our user study to investigate the important factors that real-world users take into account to determine the political stance of a news articles.

\vspace{1mm}
\noindent
\textbf{Setup.}
We have recruited 136 respondents in total from Amazon Mechanical Turk\footnote{\url{https://www.mturk.com/}}.
For a fair and reliable user study,
we chose the respondents considering a variety of aspects such as 
gender (male/female), age (from below 20 to above 50), 
education (from high school or less, to university graduation and above), 
ethnicity (e.g., Caucasian, African American, Asian, Hispanic/Latino, and Others)
and political stance (very liberal, somewhat liberal, neutral, somewhat conservative, very conservative).
We selected six news articles with different political stances and thirteen political-related factors. 
Regarding the topic of a news article,
we considered three different news topics which are highly related to political stances (i.e., Health, Environment, and Tax)\footnote{The six news articles are available at {\codeurl}.}.
For political-related factors,
we carefully chose thirteen factors in total, which have been studied in~\cite{lin2006side,gottopati2013learning,gentzkow2010drives,kuttschreuter2011framing,irvine1992rethinking},
such as context/tone, keywords, person names, topic/issue, title, images of a news article, slang words used in an article, and social/religious factors.

\vspace{1mm}
\noindent
\textbf{User study protocol.}
Then, we (1) provided the six news articles (the title, body, and image) and the thirteen carefully-chosen factors to the respondents,
(2) asked them to respond how important each factor is in their decision with a scale [1:(not at all) - 5:(very much)],
and (3) assessed each factor for political stance identification by averaging the scores rated by the respondents.

\begin{table}[h]
\centering
\caption{Result of the user study: the top-5 factors in political stance predictions and their importance scores.}
\label{table:userstudy}
\setlength\tabcolsep{8pt}
\begin{tabular}{c|cc}
\toprule
Rank & Factor name & Importance score (1-5)  \\
\midrule
1  & Context & $4.19 \pm 0.94$ \\
2  & Keyword & $4.01 \pm 0.88$ \\ 
3  & Person & $3.94 \pm 0.96$ \\ 
4  & Tone & $3.93 \pm 1.13$ \\ 
5  & Freq. used word & $3.35 \pm 1.07$ \\
\bottomrule
\end{tabular}
\end{table}

\noindent
\textbf{Result and analysis.}
Table~\ref{table:userstudy} shows the top-5 important factors in political stance prediction and their scores.
This result shows that the "context" of a new article is the most important factor in deciding its political stance, followed by keyword, person, tone, and frequently used word.
This user study result implies that
it is crucial (1) to learn the relationships among words/sentences to capture the context and tone of a news article, which is \textit{implicitly} reflected in a news article, 
and (2) to understand the interpretation and sentiment to real-world entities (e.g., keyword and person), 
which \textit{explicitly} appear in a news article.

\subsection{Data Construction}\label{sec:appendix-data}
In this section, 
we describe the process of data construction for our datasets: 
(1) a large-scale political news datasets (\textsf{AllSides-L}) 
and (2) two political knowledge graphs (\textsf{KG-lib} and \textsf{KG-con}).
All the datasets are available at: {\codeurl}.

\vspace{1mm}
\noindent
\textbf{\textsf{AllSides-L}.}
As explained in~\ref{sec:eval-setup},
we constructed a large-scale political news dataset, 
\textsf{AllSides-L}.
We collected 719,256 articles with 5 classes (left, lean left, center, lean right, and right) from Allsides.com~\footnote{\url{https://www.allsides.com/}},
which is an American website to alleviate side effects by media bias and misinformation.
Allsides.com provides political-related news articles with diverse political stances, 
where it decides the political class (e.g., left or right) of each news article based on its news outlet (e.g., CNN or Fox). 
For the political label decision, 
Allsides.com uses the three-step process:
Each news outlet is labeled by (1) domain experts, 
(2) user studies by average people with diverse political stances, 
and (3) majority voting by others who do not engage in the user studies. 
Thanks to its careful labeling, the political stances of news articles by Allsides.com are generally used as the ground truth~\cite{feng2021kgap,zhang2022kcd}. 
For the more reliability of \textsf{AllSides-L},
we consider only the news articles from the outlets receiving high scores (7-out-of-8 or better) on their labels at majority voting.

\vspace{1mm}
\noindent
\textbf{\textsf{KG-lib} and \textsf{KG-con}.}
We constructed two different political knowledge graphs (KGs), \textsf{KG-lib} and \textsf{KG-con}, via a three-step process:
(1) data collection, (2) entity/relation extraction, and (3) data cleansing.
We first collected 219,915 posts and 276,156 posts from the U.S. liberal and conservative communities, respectively (496,071 posts in total).
Since the raw posts could include many political-unrelated entities, 
we need to extract political entities and their relations from the raw posts.
To this end, we extracted 18 political-related entities and their relations, using a state-of-the-art NER (Named Entity Recognition) method~\cite{vychegzhanin2019comparison}.
Via this step, each data point is represented as a triplet: <head entity, relation, tail entity>.
For more reliability of the political knowledge graphs, 
we manually remove noises from the extracted triplets.
Finally, we constructed the two political knowledge graphs, \textsf{KG-lib} (5,581 entities and 29,967 relations) and \textsf{KG-con} (6,316 entities and 33,207 relations).

\vspace{1mm}
\noindent
\textbf{Quality of the political knowledge graphs.}
We also evaluate the quality of our political knowledge graphs.
As explained in Section~\ref{sec:proposed-ke},
we use three recent knowledge embedding methods to {\ke}: RotatE~\cite{sun2018rotate}, ModE~\cite{zhang2020learning}, and HAKE~\cite{zhang2020learning}.
We apply each knowledge embedding method to the two political knowledge graphs (i.e., \textsf{KG-lib} and \textsf{KG-con}) with varying the embedding dimensionality ($d=128,256,512$), 
and measure the quality of knowledge graphs by using five knowledge graph completion metrics: 
MR (mean rank), MMR (mean reciprocal rank), HITS@1, HITS@3, and HITS@10.
Tables~\ref{table:appendix-ke-rotate}, \ref{table:appendix-ke-mode}, and \ref{table:appendix-ke-hake} show the results.
The quality of the political knowledge embedding tends to be improved as the dimensionality of embedding increases.
Note that the quality of political knowledge embeddings could be improved in two aspects:
(1) extending the scale of political knowledge graphs (KG) 
and (2) developing a new knowledge embedding method specialized in the political stance prediction.
In future work, we plan to extend the scale of political knowledge graphs and study to design a new model architecture, specialized in capturing the relations among political entities.

\begin{table}[h]
\centering
\caption{The knowledge graph (KG) completion accuracy of RotatE~\cite{sun2018rotate} on \textsf{KG-lib} and \textsf{KG-con}.}
\label{table:appendix-ke-rotate}
\setlength\tabcolsep{2.7pt}
\begin{tabular}{l||ccc|ccc}
\toprule
\multicolumn{7}{c}{\textbf{RotatE}} \\
\midrule
\midrule
\multirow{2}{*}{Metric} & \multicolumn{3}{c}{\textbf{\textsf{KG-lib}}}  & \multicolumn{3}{c}{\textbf{\textsf{KG-con}}} \\ 
\cmidrule(lr){2-4} \cmidrule(lr){5-7} 
& $d=128$ & $d=256$ & $d=512$ & $d=128$ & $d=256$ & $d=512$ \\ 
\midrule
MR & 632.69 & 573.84 & 567.85 & 728.78 & 654.26 & 640.45\\
MRR & 0.1312 & 0.1700 & 0.1859 & 0.1079 & 0.1494 & 0.1633 \\
HITS@1 & 0.0842 & 0.1089 & 0.1209 & 0.0692 & 0.0974 & 0.1093 \\
HITS@3 & 0.1316 & 0.1801 & 0.1985 & 0.1059 & 0.1549 & 0.1693 \\
HITS@10 & 0.2133 & 0.2859 & 0.3148 & 0.1743 & 0.2429 & 0.2625 \\
\bottomrule
\end{tabular}
\end{table}

\begin{table}[h]
\centering
\caption{The knowledge graph (KG) completion accuracy of ModE~\cite{zhang2020learning} on \textsf{KG-lib} and \textsf{KG-con}.}
\label{table:appendix-ke-mode}
\setlength\tabcolsep{2.7pt}
\begin{tabular}{l||ccc|ccc}
\toprule
\multicolumn{7}{c}{\textbf{ModE}} \\
\midrule
\midrule
\multirow{2}{*}{Metric} & \multicolumn{3}{c}{\textbf{\textsf{KG-lib}}}  & \multicolumn{3}{c}{\textbf{\textsf{KG-con}}} \\ 
\cmidrule(lr){2-4} \cmidrule(lr){5-7} 
& $d=128$ & $d=256$ & $d=512$ & $d=128$ & $d=256$ & $d=512$ \\ 
\midrule
MR & 690.14 & 622.40 & 645.23 & 777.11 & 740.88 & 723.90 \\
MRR & 0.1312 & 0.1700 & 0.1859 & 0.1128 & 0.1354 & 0.1501 \\
HITS@1 & 0.0842 & 0.1089 & 0.1209 & 0.0685 & 0.0801 & 0.0913 \\
HITS@3 & 0.1316 & 0.1801 & 0.1985 & 0.1127 & 0.1404 & 0.1567 \\
HITS@10 & 0.2133 & 0.2859 & 0.3148 & 0.1981 & 0.2458 & 0.2648 \\
\bottomrule
\end{tabular}
\end{table}

\begin{table}[h]
\centering
\caption{The knowledge graph (KG) completion accuracy of HAKE~\cite{zhang2020learning} on \textsf{KG-lib} and \textsf{KG-con}.}
\label{table:appendix-ke-hake}
\setlength\tabcolsep{2.7pt}
\begin{tabular}{l||ccc|ccc}
\toprule
\multicolumn{7}{c}{\textbf{HAKE}} \\
\midrule
\midrule
\multirow{2}{*}{Metric} & \multicolumn{3}{c}{\textbf{\textsf{KG-lib}}}  & \multicolumn{3}{c}{\textbf{\textsf{KG-con}}} \\ 
\cmidrule(lr){2-4} \cmidrule(lr){5-7} 
& $d=128$ & $d=256$ & $d=512$ & $d=128$ & $d=256$ & $d=512$ \\ 
\midrule
MR & 593.76 & 597.58 & 606.03 & 694.30 & 684.55 & 685.92 \\
MRR & 0.1474 & 0.1688 & 0.1787 & 0.1311 & 0.1498 & 0.1639 \\
HITS@1 & 0.0904 & 0.1102 & 0.1167 & 0.0831 & 0.0992 & 0.1120 \\
HITS@3 & 0.1541 & 0.1761 & 0.1895 & 0.1348 & 0.1550 & 0.1704 \\
HITS@10 & 0.2595 & 0.2844 & 0.3013 & 0.2205 & 0.2434 & 0.2612 \\
\bottomrule
\end{tabular}
\end{table}

\subsection{Reliability of Experiments}\label{sec:appendix-additional-exp}
In this section, we verify the reliability of the experimental results, used in our empirical evaluation (EQ1. Accuracy in Section~\ref{sec:eval-result}).

\vspace{1mm}
\noindent
\textbf{Evaluation protocol.}
As mentioned in Section~\ref{sec:eval-result},
we compare the model accuracy of {\m} with the experimental results of the seven baseline methods, 
which have been reported in~\cite{zhang2022kcd}, on the two widely used datasets (i.e., \textsf{SemEval} and \textsf{AllSides-S}).
To evaluate the reliability of the accuracy of {\m}, 
we (1) implement the five baseline language models, using their available codes, with a softmax layer for the final political stance prediction, 
(2) perform the baseline methods on the SemEval and AllSides-S datasets,
(3) measure their political stance prediction accuracies,
and (4) compare the results (i.e., Validation Acc.) with the reported results (i.e., Reported Acc.).
For KGAP~\cite{feng2021kgap} and KCD~\cite{zhang2022kcd}, however,
we cannot obtain the results because some parts of their source codes are not provided.
More specifically, the source code to generate the cue embeddings for KCD and the source code to generate the knowledge embeddings for KGAP are missing, respectively.\footnote{KGAP: \url{https://github.com/BunsenFeng/news_stance_detection}, KCD: \url{https://github.com/Wenqian-Zhang/KCD}}.

\vspace{1mm}
\noindent
\textbf{Result and analysis.}
Tables~\ref{table:appendix-model-accuracy-semeval} and~\ref{table:appendix-model-accuracy-allsides} show the averaged accuracy and standard deviation of each method and the reported results~\cite{zhang2022kcd} on \textsf{SemEval} and \textsf{AllSides-S}.
The results that we obtain (i.e., Validation Acc.) are quite similar as (sometimes higher than) those reported in~\cite{zhang2022kcd} (i.e., Reported Acc.).
Based on these results, 
we believe that our evaluation protocol and experimental results could be justified and reliable.
As a result, considering that {\m} consistently outperforms all baseline methods with a very low standard deviation,
our experimental results demonstrate the superiority of {\m} over existing political stance prediction methods.

\begin{table}[h]
\centering
\caption{Comparison of our experimental results with the reported results~\cite{zhang2022kcd} on \textsf{SemEval} (The bold font indicates the results better than the reported results).}
\label{table:appendix-model-accuracy-semeval}
\setlength\tabcolsep{12pt}
\begin{tabular}{c||c|c}
\toprule
\multirow{2}{*}{Method} & \multicolumn{2}{c}{\textbf{\textsf{SemEval}}} \\ 
\cmidrule(lr){2-3}
 & Validation Acc. & Reported Acc. \\ 
\midrule
\midrule
\textbf{Word2Vec}  & \textbf{0.7076} $\pm$ 0.0104 & 0.7027  \\
\textbf{GloVe}     & \textbf{0.8077} $\pm$ 0.0251 & 0.8071 \\
\textbf{ELMo}      & 0.8666 $\pm$ 0.0197 & 0.8678 \\
\textbf{BERT}      & \textbf{0.8769} $\pm$ 0.0156 & 0.8692 \\
\textbf{RoBERTa}   & \textbf{0.8923} $\pm$ 0.0112 & 0.8708 \\
\midrule

\textbf{KGAP}      & N/A & 0.8956 \\
\textbf{KCD}       & N/A & 0.9087 \\
\midrule

\textbf{{\m}}-RotatE    & 0.9426 $\pm$ 0.0258 & N/A \\
\textbf{{\m}}-HAKE      & 0.9395 $\pm$ 0.0290 & N/A \\
\textbf{{\m}}-ModE      & 0.9521 $\pm$ 0.0183 & N/A \\
\bottomrule
\end{tabular}
\end{table}

\begin{table}[h]
\centering
\caption{Comparison of our experimental results with the reported results~\cite{zhang2022kcd} on \textsf{AllSides-S} (The bold font indicates the results better than the reported results).}
\label{table:appendix-model-accuracy-allsides}
\setlength\tabcolsep{12pt}
\begin{tabular}{c||c|c}
\toprule
\multirow{2}{*}{Method} & \multicolumn{2}{c}{\textbf{\textsf{AllSides-S}}} \\ 
\cmidrule(lr){2-3}
 & Validation Acc. & Reported Acc. \\ 
\midrule
\midrule
\textbf{Word2Vec}  & \textbf{0.4977} $\pm$ 0.0082 & 0.4858 \\
\textbf{GloVe}     & 0.6978 $\pm$ 0.0204 & 0.7101 \\
\textbf{ELMo}      & 0.8085 $\pm$ 0.0178 & 0.8197 \\
\textbf{BERT}      & 0.8201 $\pm$ 0.0101 & 0.8246 \\
\textbf{RoBERTa}   & \textbf{0.8682} $\pm$ 0.0081 & 0.8535 \\
\midrule

\textbf{KGAP}      & N/A & 0.8602 \\
\textbf{KCD}       & N/A & 0.8738 \\
\midrule

\textbf{{\m}}-RotatE    & 0.9151 $\pm$ 0.0105 & N/A \\
\textbf{{\m}}-HAKE      & 0.9216 $\pm$ 0.0041 & N/A \\
\textbf{{\m}}-ModE      & 0.9256 $\pm$ 0.0098 & N/A \\
\bottomrule
\end{tabular}
\end{table}

\end{document}